\documentclass[times,twocolumn,final,authoryear]{elsarticle}

\usepackage{style}
\usepackage{framed,multirow}
\usepackage{amssymb}
\usepackage{latexsym}
\usepackage{url}
\usepackage[table]{xcolor}
\usepackage{xcolor}
\usepackage{framed,multirow}

\usepackage{amssymb}
\usepackage{latexsym}
\usepackage{comment}

\begin{document}

\setcounter{page}{1}

\begin{frontmatter}

\title{Exploiting Multimodal Synthetic Data for Egocentric Human-Object Interaction Detection in an Industrial Scenario}

\author[1]{Rosario Leonardi\corref{cor1}} 
\cortext[cor1]{Corresponding author.}
\ead{rosario.leonardi@phd.unict.com}
\author[1,2]{Francesco Ragusa}
\author[1,2]{Antonino Furnari}
\author[1,2]{Giovanni Maria Farinella}

\address[1]{FPV@IPLAB, Department of Mathematics and Computer Science - University of Catania, Catania 95125, Italy}
\address[2]{Next Vision s.r.l, Spinoff of the University of Catania - Viale Andrea Doria 6, Catania 95125, Italy}

\begin{abstract}
In this paper, we tackle the problem of Egocentric Human-Object Interaction (EHOI) detection in an industrial setting. To overcome the lack of public datasets in this context, we propose a pipeline and a tool for generating synthetic images of EHOIs paired with several annotations and data signals (e.g., depth maps or segmentation masks). Using the proposed pipeline, we present \textit{EgoISM-HOI} a new multimodal dataset composed of synthetic EHOI images in an industrial environment with rich annotations of hands and objects. To demonstrate the utility and effectiveness of synthetic EHOI data produced by the proposed tool, we designed a new method that predicts and combines different multimodal signals to detect EHOIs in RGB images. Our study shows that exploiting synthetic data to pre-train the proposed method significantly improves performance when tested on real-world data. Moreover, to fully understand the usefulness of our method, we conducted an in-depth analysis in which we compared and highlighted the superiority of the proposed approach over different state-of-the-art class-agnostic methods. To support research in this field, we publicly release the datasets, source code, and pre-trained models at \url{https://iplab.dmi.unict.it/egoism-hoi}.
\end{abstract}

\end{frontmatter}


\section{Introduction}\label{sec:introduction}
In recent years, wearable devices have become increasingly popular as they offer a first-person perspective of how users interact with the world around them. One of the advantages of wearable devices is that they allow the collection and processing of visual information without requiring users to hold any devices with their hands, enabling them to perform their activities in a natural way. Intelligent systems can analyze this visual information to provide services to support humans in different domains such as activities of daily living \citep{damen2014you, Damen2018ScalingEV, Grauman2021Ego4DAT}, cultural sites \citep{farinella2019vedi} and industrial scenarios \citep{sener2022assembly101, Mazzamuto_VISAPP_2023_Wearable_Device_CAMERA_READY}. In particular, egocentric vision can be adopted in the industrial context to understand workers’ behavior, improve workplace safety, and increase overall productivity. For example, by detecting the hands of the workers and determining which objects they are interacting with, it is possible to monitor object usage, provide information on the procedures to be carried out, and improve the safety of workers by issuing reminders when dangerous objects are manipulated.

Previous works have investigated the problem of Human-Object Interaction detection (HOI) considering either third-person \citep{GkioxariDRHOI, Liao2020PPDMPP} or first-person \citep{liu2022hoi4d, zhang2022fine} points of view. While these works have considered generic scenarios (e.g., COCO objects) or class-agnostic settings \citep{Shan2020UnderstandingHH}, their use in industrial contexts is still understudied due to the limited availability of public datasets \citep{Ragusa2021TheMD, sener2022assembly101}. To develop a system capable of detecting Egocentric Human-Object Interactions (EHOI) in this context, it is generally required to collect and label large amounts of domain-specific data, which could be expensive in terms of costs and time and not always possible due to privacy constraints and industrial secrets \citep{Ragusa2021TheMD}. 
\begin{figure*}[t]
    \centering
    \includegraphics[scale=1]{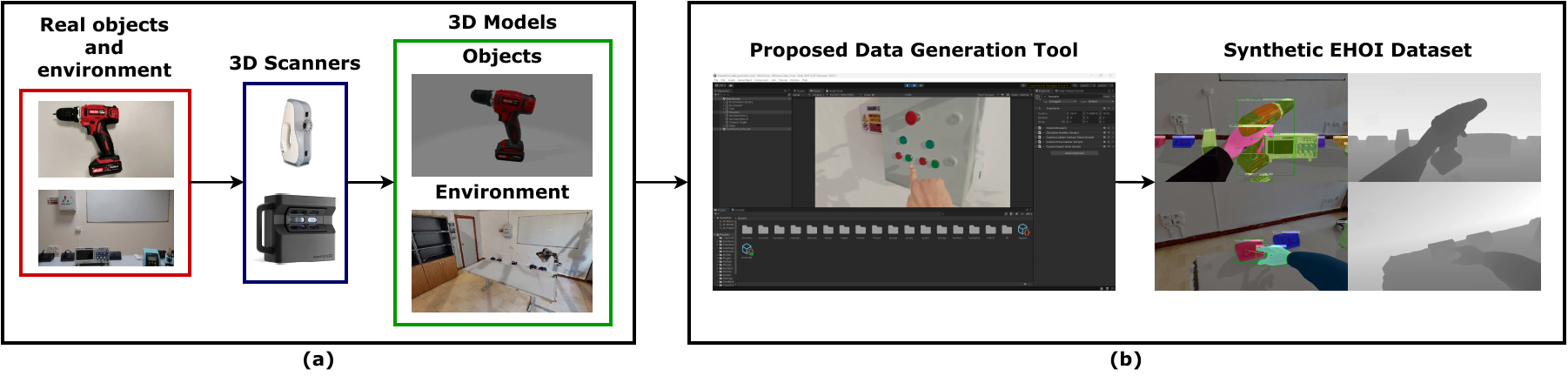}
    \caption{Synthetic EHOI images generation pipeline. (a) We use 3D scanners to acquire 3D models of the objects and environment. (b) We hence use the proposed data generation tool to create the synthetic dataset.}
    \label{fig:data_generation_pipeline}
\end{figure*}

In this paper, we investigate whether the use of synthetic data in first-person vision can mitigate the need for labeled real domain-specific data in model training, which would greatly reduce the cost of gathering a suitable dataset for model development. We propose a pipeline (see Fig.~\ref{fig:data_generation_pipeline}) and a tool that, leveraging 3D models of the target environment and objects, produces a large number of synthetic EHOI image examples, automatically labeled with several annotations, such as hand-object 2D-3D bounding boxes, object categories, hand information (i.e., hand side, contact state, and associated active objects) as well as multimodal signals such as depth maps and instance segmentation masks.

Exploiting the proposed pipeline, we present \textit{EgoISM-HOI} (Egocentric Industrial Synthetic Multimodal dataset for Human-Object Interaction detection), a new photo-realistic dataset of EHOIs in an industrial scenario with rich annotations of hands, objects, and active objects (i.e., the objects the user is interacting with), including class labels, depth maps, and instance segmentation masks (see Fig.~\ref{fig:data_generation_pipeline} (b)). To assess the suitability of the synthetic data generated with the proposed protocol to tackle the EHOI detection task on target real data, we further acquired and labeled 42 real egocentric videos in an industrial laboratory in which different subjects perform test and repair operations on electrical boards\footnote{Note that both real and synthetic data were acquired in the same environment and with the same objects}. We annotated all EHOIs instances of the images identifying the frames in which interactions occur and all active objects with a bounding box associated with the related object class. In addition, we labeled the hands and all the objects in the images.

We investigated the potential of using the generated synthetic multimodal data, including depth maps and instance segmentation masks, to improve the performance of EHOI detection methods. Specifically, we designed an EHOI detection approach based on the method proposed in \citet{Shan2020UnderstandingHH} which makes use of the different multimodal signals available within our dataset. Experiments show that the proposed method outperforms baseline approaches based on the exploitation of class-agnostic models trained on out-of-domain real-world data. Indeed, the proposed method achieves good performance when trained with our synthetic data and a very small amount of real-world data. Additional experiments show that, by leveraging multimodal signals, the accuracy and robustness of our EHOI detection system increased.

The contributions of this study are the following: 1) we propose a pipeline that exploits 3D models of real objects and environments to generate thousands of domain-specific synthetic egocentric human-object interaction images paired with several labels and modalities; 2) we present \textit{EgoISM-HOI}, a new multimodal dataset of synthetic EHOIs in an industrial scenario with rich annotations of hands and objects. To test the ability of models to generalize to real-world data, we acquire and manually labeled real-world images of EHOIs in the target environment; 3) we design a new method for EHOI detection that exploits additional modalities, such as depth maps and instance segmentation maps to enhance the performance of classic HOI detection approaches; 4) we perform extensive evaluations to highlight the benefit of using synthetic data to pre-train EHOI detection methods, mainly when a limited set of real data is available, and report improvements of our approach over classical class-agnostic state-of-the-art methods; 5) we release the dataset and code publicly at the following link: \url{https://iplab.dmi.unict.it/egoism-hoi}.

The remainder of this paper is organized as follows. Section~\ref{sec:related_work} provides a detailed summary of the related work. Section~\ref{sec:proposed_ehoi_generation_pipeline} details the proposed data generation pipeline. Section~\ref{sec:egoism_hoi} describes the proposed dataset. Section~\ref{sec:approach} introduces our multimodal EHOI detection method. Section~\ref{sec:experimental_results} reports and discusses the performed experiments and ablation studies. Finally, Section~\ref{sec:conclusion} concludes the paper.


\section{Related Work}\label{sec:related_work}
In this Section, we discuss datasets and state-of-the-art methods for detecting human-object interactions from images and videos acquired from both third~(TPV) and first-person vision~(FPV).

\subsection{Datasets for Human-Object Interaction Detection}
Previous works have proposed benchmark datasets to study human-object interactions from a third-person vision. The datasets, such as \textit{PASCAL VOC} \citep{everingham2009pascal}, \textit{V-COCO} \citep{Gupta2015VisualSR}, \textit{HICO} \citep{Chao2015HICOAB}, \textit{HICO-DET} \citep{chao:wacv2018}, \textit{AmbiguousHOI} \citep{li2020detailed}, \textit{HOI-A} \citep{Liao2020PPDMPP}, and \textit{BEHAVE} \citep{bhatnagar22behave}, offer diverse annotations and cover a wide range of scenarios. Most related to our study is \textit{100 Days of Hands} \citep{Shan2020UnderstandingHH} which is a large-scale dataset of human-object interactions containing more than 131 days of video footage acquired from both third and first-person points of view. The authors extracted 100K frames and annotated with bounding boxes 189.6K hands and 110.1K objects involved in interactions. Moreover, for each hand, they annotated the contact state considering five different classes (i.e., \textit{none, self, other-person, non-portable object}, and \textit{portable object}).
Differently from previous works, our study focuses on understanding human-object interactions from a first-person point of view with the exploitation of synthetic generated data.

Owing to the aforementioned vantage point given by wearable cameras, previous works have proposed datasets to study human-object interactions from first-person vision. \textit{EgoHands} \citep{Bambach2015LendingAH} is a dataset composed of egocentric video pairs of people interacting with their hands in different daily-life contexts, where they are involved in four social situations (i.e., playing cards, playing chess, solving puzzles, and playing Jenga). It is composed of 130,000 frames and 4,800 pixel-level segmentation masks of hands. \textit{EPIC-KITCHENS-100} \citep{Damen2021RESCALING} contains over 100 hours, 20 million frames, and 90,000 actions in 700 variable-length videos of unscripted activities in 45 kitchen environments. The authors provide spatial annotations of (1) instance segmentations masks using Mask R-CNN \citep{mask_rcnn} and (2) hand and active object bounding boxes labeled with the system introduced in \citet{Shan2020UnderstandingHH}. \citet{VISOR2022} proposed \textit{VISOR}, an extension of \textit{EPIC-KITCHENS-100}, which comprises pixel annotations and a benchmark suite for segmenting hands and active objects in egocentric videos. It contains 272,000 manual segmented semantic masks of 257 object classes, 9.9 million interpolated dense masks, and 67,000 hand-object relations. \textit{EGTEA Gaze+} \citep{li2020eye} contains more than 28 hours of egocentric video acquired by subjects performing different meal preparation tasks. The authors provide several annotations, including binocular gaze tracking data, frame-level action annotations, and 15K hand segmentation masks. Recognizing EHOIs could be particularly useful in industrial scenarios, for example, to optimize production processes or to increase workplace safety. \textit{MECCANO} \citep{Ragusa2021TheMD, ragusa2022meccano} is a multimodal dataset of FPV videos for human behavior understanding collected in an industrial-like scenario. It includes gaze signals, depth maps, and several annotations. MECCANO has been explicitly annotated to study EHOIs with bounding boxes around the hands and active objects, and verbs that describe the interactions. \textit{Assembly101} \citep{sener2022assembly101} is a multi-view action dataset of people assembling and disassembling 101 toy vehicles. It contains 4321 video sequences acquired simultaneously from 8 TPV and 4 FPV cameras, 1M fine-grained action segments, and 18 million 3D hand poses. \textit{Ego4D} \citep{Grauman2021Ego4DAT} is a multimodal video dataset to study egocentric perception. The dataset contains more than 3,500 video hours of daily life activity captured by 931 subjects and additional modalities such as eye gaze data, audio, and 3D mesh of environments. EGO4D has been annotated with bounding boxes around the hands and objects involved in the interactions. \textit{HOI4D} \citep{liu2022hoi4d} is a large-scale 4D egocentric dataset for human-object interaction detection. \textit{HOI4D} contains more than 2 million RGB-D egocentric video frames in different indoor environments of people interacting with 800 object instances. 

Unlike these works, we aim to study the usefulness of synthetic data for training models which need to be deployed in a specific environment. To this aim, we provide \textit{EgoISM-HOI}, a photo-realistic multimodal dataset of synthetic images for understanding human-object interactions acquired in an industrial scenario, paired with labeled real-world images of egocentric human-object interactions in the same target environment. Our dataset contains RGB-D images and rich automatically labeled annotations of hands, objects, and active objects, including bounding boxes, object categories, instance segmentation masks, and interaction information (i.e., hand contact state, hand side, and hand-active object relationships).

\subsection{Human-Object Interaction simulators and synthetic datasets}
This line of research focused on providing 3D simulators which are able to generate automatically labeled synthetic data \citep{kolve2017ai2, savva2019habitat, xia2020interactive, Hwang2020ElderSimAS, quattrocchi2023Outline_SAFER}. While these tools allow simulating an agent that navigates in an indoor environment, there are fewer choices for simulating object interaction. \citet{Mueller_real_time_hand_tracking_2017} proposed a data generation framework that tracks and combines real human hands with virtual objects to generate photorealistic images of hand-object interactions. Using the proposed tool, the authors introduced \textit{SynthHands}, a dataset that contains around 200K RGB-D images of hand-object interactions acquired from 5 FPV virtual cameras. \textit{ManipulaTHOR} \citep{ehsani2021manipulathor} is an extension of the \textit{AI2-THOR} framework \citep{kolve2017ai2} that adds a robotic arm to virtual agents, enabling the interaction with objects. Thanks to this framework, the authors introduced the \textit{Arm POINTNAV} dataset, which contains interactions in 30 kitchen scenes, 150 object categories, and 12 graspable object categories. \citet{HassonLJR} introduced the \textit{ObMan} dataset, a large-scale synthetic image dataset of hand-object interactions. The peculiarity of this work is that the authors used the \textit{GraspIt} software \citep{miller2004graspit} to improve the photo-realism of the generated interactions. The generated dataset contains more than 20,000 hand-object interactions in which the background is randomized by choosing images from the \textit{LSUN} \citep{yu2015lsun} and \textit{ImageNet} \citep{russakovsky2015imagenet} datasets. \citet{wang2022dexgraspnet} introduced \textit{DexGraspNet}, a large-scale synthetic dataset for robotic dexterous grasping containing 1.32M grasps of 5355 objects among 133 object categories. \citet{ye2023affordance} proposed an approach for synthesizing virtual human hands interacting with real-world objects from RGB images. 

Differently from these works, our generation pipeline has been specifically designed to obtain accurate 3D reconstructions of a target environment and the objects it contains. 3D models of the target environment and objects are used by our tool to generate realistic egocentric hand-object interactions that integrate coherently with the surrounding environment. Moreover, our tool allows the customization of several parameters of the virtual scene, for example, by randomizing the light points, the position of the virtual object in the environment, or the virtual agent's clothing. In addition, the proposed tool is able to output several annotations automatically labeled and data signals, such as 2D-3D bounding boxes, hand labels (i.e., hand contact state and hand side), instance segmentation masks, and depth maps. Another difference with respect to the aforementioned works is that our tool is designed to automatically generate interactions from a first-person point of view without using any additional real-world data or specific hardware devices other than 3D models.

\subsection{Methods for Detecting Human-Object Interactions}
In the past years, the human-object interaction detection task has been studied from the third-person point of view \citep{Gupta2015VisualSR, Chao2015HICOAB, chao:wacv2018}. \citet{GkioxariDRHOI} proposed a method for detecting human-object interactions in the form of \textit{$<$human, verb, object$>$} triplets, where bounding boxes around objects and humans are also predicted. Specifically, they extended the state-of-the-art object detector Faster R-CNN \citep{faster_rccn} with an additional human-centric branch that uses the features extracted by the backbone to predict a score for candidate human-object pairs and an action class. \citet{Liao2020PPDMPP} proposed a method called \textit{PPDM} (Parallel Point Detection and Matching) that defines an HOI as a triplet \textit{$<$human point, interaction point, object point$>$} composed of three points associated with the human, the active object and the interaction location. Recently, several works figured out the HOI detection task by proposing transformer-based models. \citet{zhang2022upt} proposed a new two-stage detector based on a transformer architecture to detect interactions. \citet{wu2022mining} proposed an approach for learning a body-part saliency map, which contains informative cues of the person involved in the interaction and other persons in the image, in order to boost HOI detection methods \citep{chao:wacv2018, gao2018ican}. \citet{FGAHOI} introduced a transformer-based human-object interaction detector that uses a multi-scale feature extractor and a multi-scale sampling strategy to predict the HOI instances from images with noisy backgrounds in the form of $<b_{h}, b_{o},c_{o},c_{v}>$ quadruplet, where $b_{h}$ and $b_{o}$ represent the human and object boxes, and $c_{o}$ and $c_{v}$ the object class and the verb class. While previous works all addressed the HOI modeling detecting a bounding box around the human, \citet{Shan2020UnderstandingHH} addressed the HOI detection task by predicting information about human hands, such as hand location, side, contact state, and, in case of an interaction, a box around the object touched by the hand. \citet{zhang2022fine} proposed to use a contact boundary, i.e. the contact region between the hand and the interacting object, to model the interaction relationship between hands and objects. \citet{Fu2021SequentialDF} designed an approach for HOI detection that introduced a new pixel-wise voting function for improving the active object bounding box estimation. \citet{BenaventHOI2022} proposed an architecture for human-object interaction detection estimation based on two YOLOv4 object detectors \citep{bochkovskiy2020yolov4} and an attention-based method. Recently, some work investigated the use of additional modalities, such as 6DOF hand poses or semantic segmentation masks, to learn more robust representations of human-object interactions. \citet{lu2021egocentric} introduced an approach that uses contextual information, i.e. hand pose, hand mask, and object mask, to improve the performance of HOI detection systems. 

In this work, we focused on detecting human-object interactions from FPV, where, in most cases, the hands are the only portion of the body visible in the images. To this aim, we designed an approach for detecting egocentric human-object interactions using different multimodal signals available within our \textit{EgoISM-HOI} dataset. Similar to \citet{Shan2020UnderstandingHH}, our method detects hands from RGB images using a two-stage object detector and predicts some attributes of the latter, such as hands side, hands contact state, and the objects involved in the interactions. Additionally, our approach is able to detect all objects present in the image and infer their category. Similar to \citet{lu2021egocentric, zhang2022fine}, we exploit multimodal signals (i.e., depth maps and hand segmentation masks) to predict the hand contact state. 

\section{Proposed EHOI Generation Pipeline}\label{sec:proposed_ehoi_generation_pipeline}
\begin{figure}[t!]
    \centering
    \includegraphics[scale=1]{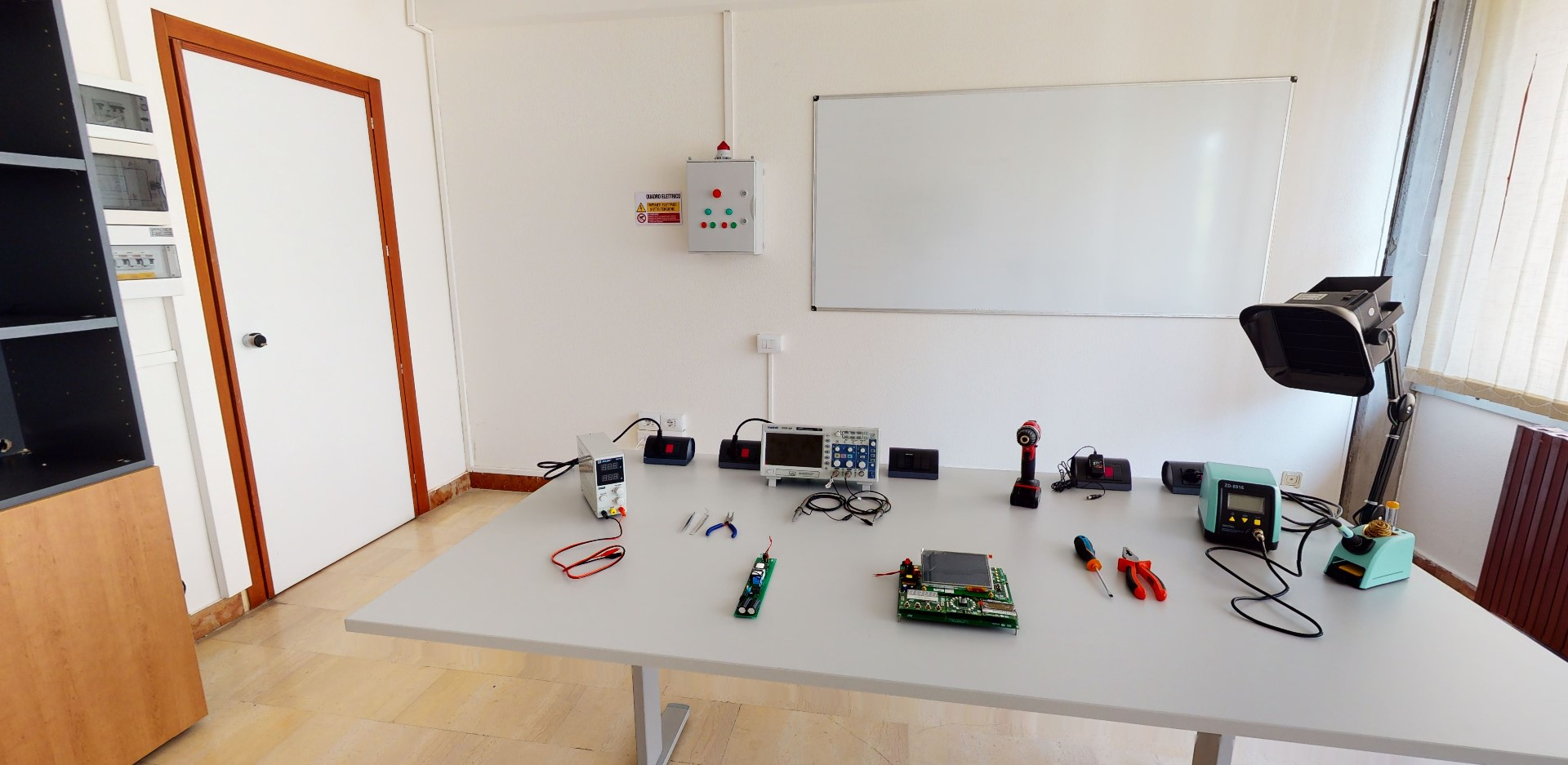}
    \caption{A picture of the ENIGMA Lab.}
    \label{fig:industrial_lab}
\end{figure}
\begin{figure}[t]
    \centering
    \includegraphics[scale=1]{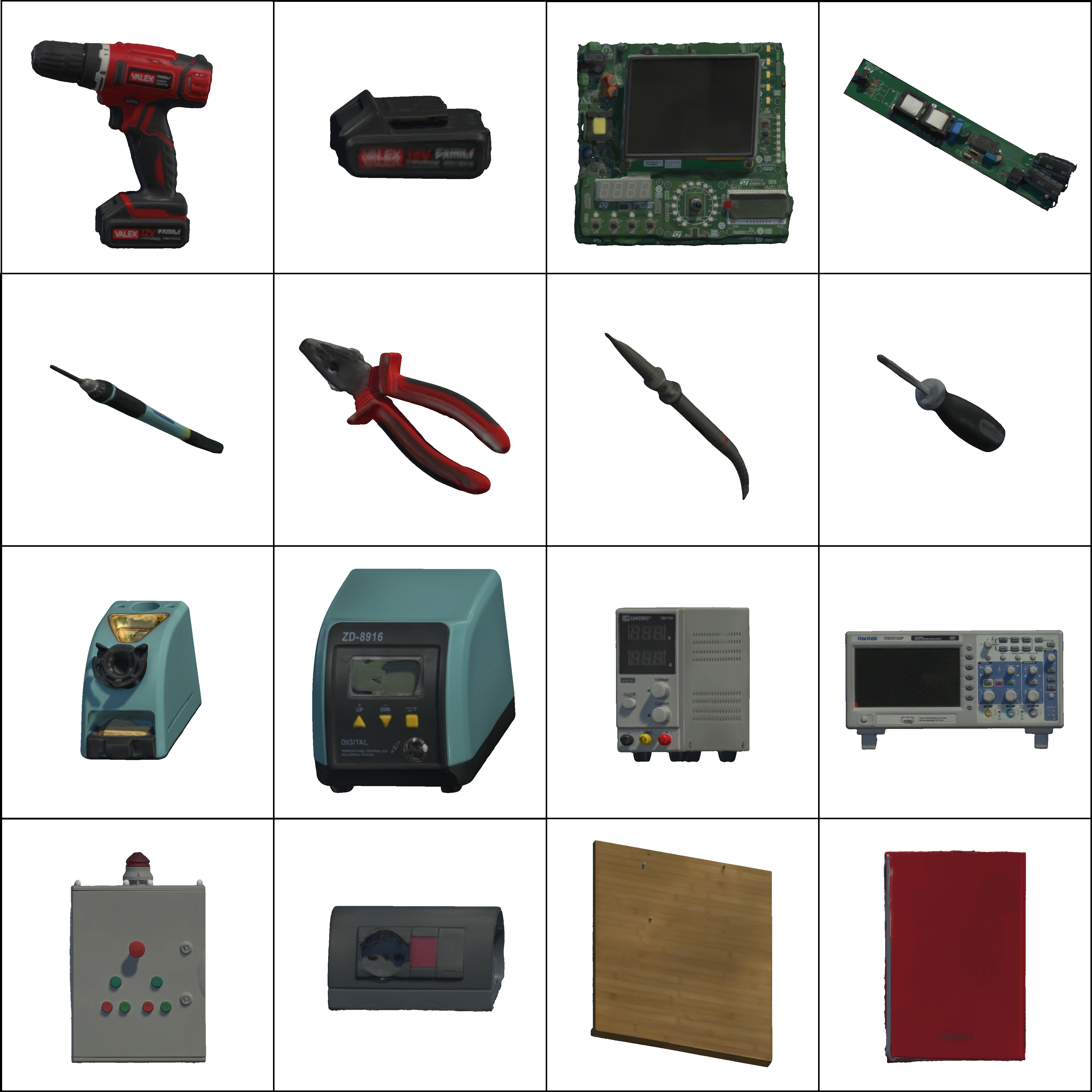}
    \caption{3D models of the 19 objects considered for the experiments.}
    \label{fig:object_models}
\end{figure}

To study the egocentric human-object interaction detection task in a realistic industrial scenario, we have set up a laboratory called \textit{ENIGMA Lab} (Figure~\ref{fig:industrial_lab}) that contains different types of work tools and equipment. Specifically, we considered the following 19 object categories: \textit{power supply, oscilloscope, welder station, electric screwdriver, screwdriver, pliers, welder probe tip, oscilloscope probe tip, low voltage board, high voltage board, register, electric screwdriver battery, working area, welder base, socket, left red button, left green button, right red button}, and \textit{right green button}. Figure~\ref{fig:object_models} shows the acquired 3D models of all the objects considered for the experiments. Note that the categories \textit{left red button, left green button, right red button}, and \textit{right green button}, refer to each button of the electrical panel shown in the bottom-left corner of Figure~\ref{fig:object_models}. 

We propose a pipeline for generating and labeling synthetic human-object interactions from a first-person point of view using 3D models of the target environment and objects, which can be cheaply collected using commercial scanners. Figure~\ref{fig:data_generation_pipeline} shows the overall scheme of our EHOI data generation pipeline, which consists of two main phases: 1) the collection of the 3D models, and 2) the generation of EHOI synthetic images using the proposed tool.

In our study, we noted that high-quality object reconstructions are necessary to generate realistic EHOIs, while high accuracy is not required for environment reconstruction. We used two different 3D scanners to create 3D models. Specifically, we used the structured-light 3D scanner \textit{Artec Eva}\footnote{\url{https://www.artec3d.com/portable-3d-scanners/artec-eva-v2}\label{foot:artec}} for scanning the objects, and a \textit{MatterPort}\footnote{\url{https://matterport.com/}\label{foot:matterport}} device for the environment.

\begin{figure}[t]
    \centering
    \includegraphics[scale=1]{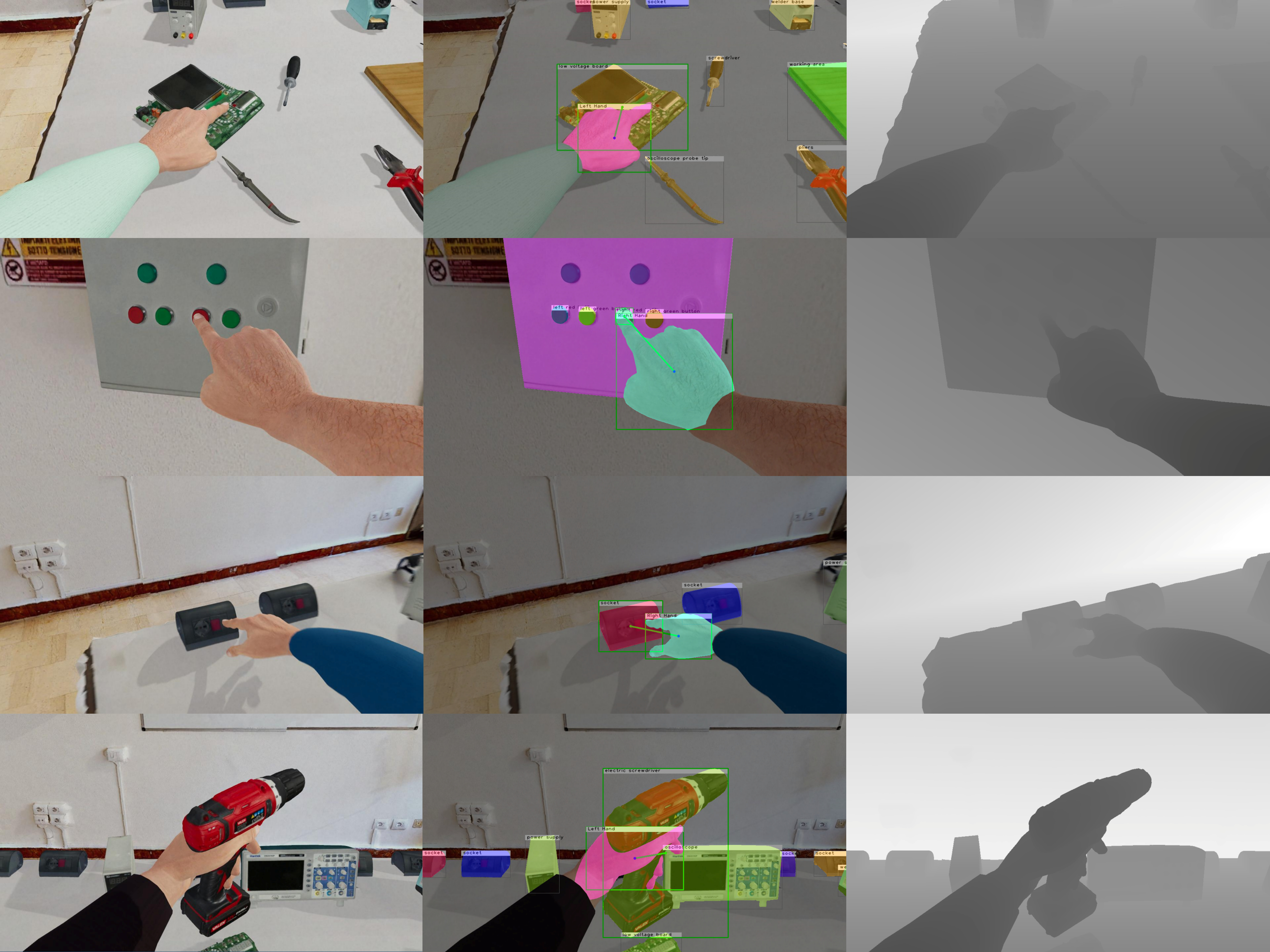}
    \caption{Examples of synthetic images (left) with the corresponding annotations (center) and depth maps (right) generated with the proposed tool.} 
    \label{fig:egoism_synth_anns}
\end{figure}

We developed a tool based on the Unity\footnote{\url{https://unity.com/}} engine which exploits 3D models of the objects and the environment to generate synthetic egocentric human-object interaction images together with the following data: 1) RGB images (see Fig.~\ref{fig:egoism_synth_anns} - left), 2) depth maps (see Fig.~\ref{fig:egoism_synth_anns} - right), 3) instance segmentation masks (see Fig.~\ref{fig:egoism_synth_anns} - center), 4) bounding boxes for hands and objects including the object categories, 5) EHOI's metadata, such as information about associations between hands and objects in contact (which hand is in contact with which object), and hand attributes (i.e., hand side, and hand contact state). Differently from our previous work \citep{leonardi2022egocentric}, the new tool streamlines the setup of virtual scenes, enhances interaction realism, facilitates the generation of diverse modalities, introduces support for RGB video generation, and allows customization of various scene parameters.

\begin{figure}[t]
    \centering
    \includegraphics[scale=1]{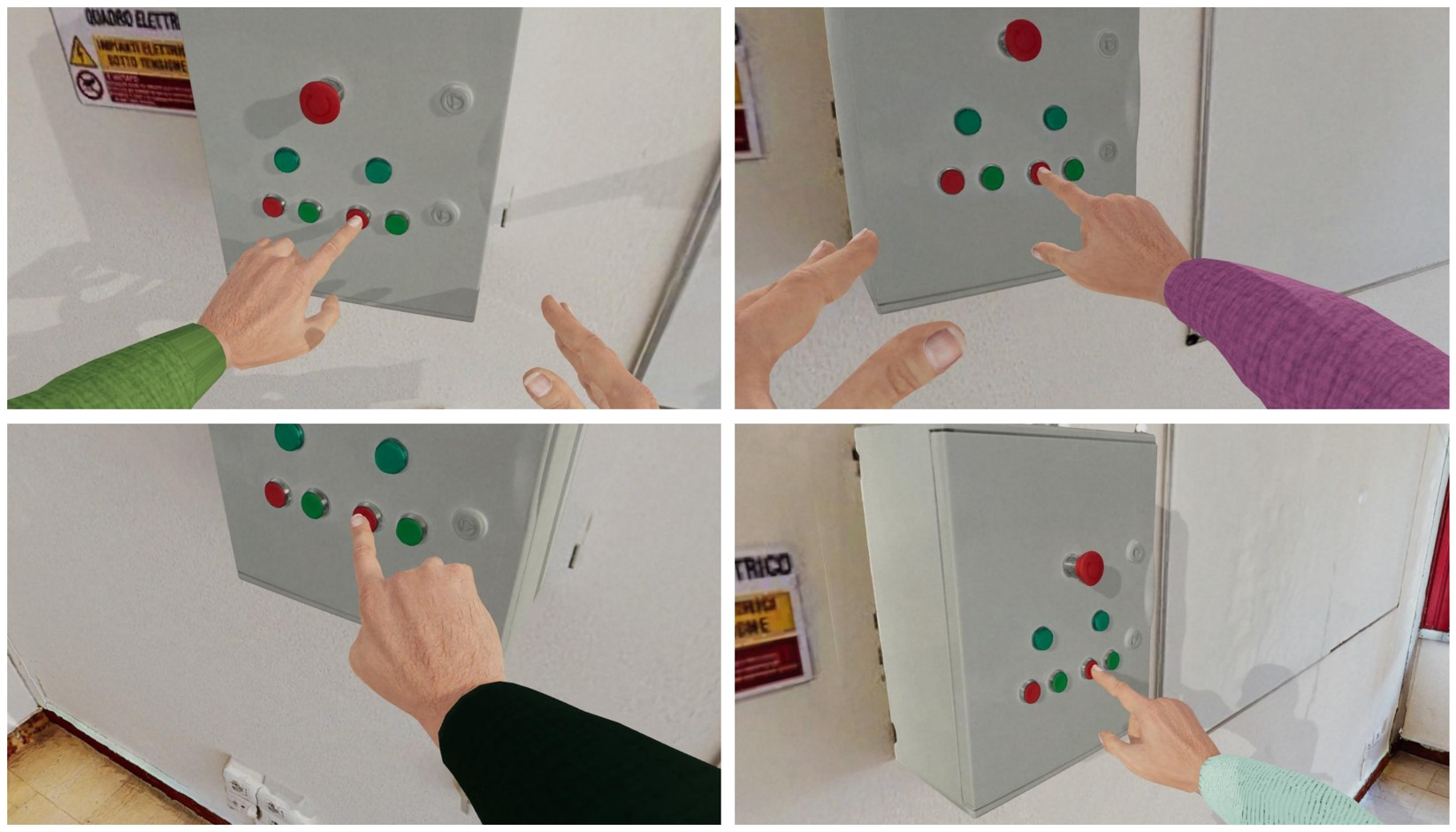}
    \caption{Our tool is able to randomize different aspects of the virtual scene, such as the camera and user positions or the shirt's texture and color.} 
    \label{fig:egoism_randomization_examples}
\end{figure}

Our system exploits the \textit{Unity Perception package} \citep{unity-perception2022}, which offers different tools for generating large-scale synthetic datasets. This package allows to randomize some aspects of the virtual scene, such as the intensity and the color of the lights, the object textures, the presence and amount of motion blur, as well as visual effects like noise, to make the virtual scene more realistic, and adds further diversity to the generated dataset, making it more representative of the real-world environment. In addition, to include different randomized aspects, we created the following randomizers:
\begin{itemize}
    \item \textit{SurfaceObjectPlacementRandomizer}: Randomizes the position of a group of objects on a flat surface;
    \item \textit{CustomRotationRandomizer}: Randomizes object rotation by respecting the constraints of each rotation axis;
    \item \textit{PlayerPlacementRandomizer}: Randomizes the location of the virtual agent in the environment;
    \item \textit{TextureShirtRandomizer}: Randomizes the texture and color of the virtual agent's shirt;
    \item \textit{CameraRandomizer}: Randomizes the observed point of the FPV camera;
\end{itemize}
Examples of randomization are shown in Figure~\ref{fig:egoism_randomization_examples}.

The Unity perception package provides a component called \textit{Scenario} which allows to control the execution flow of the simulation by setting standard simulation parameters, such as the number of iterations, the seed of the randomizers, and the number of frames to acquire for each iteration. We have extended the basic \textit{Scenario} by adding the following parameters: 1) the probability that an interaction will occur in the current iteration, 2) the target object with which the virtual agent will interact in the current interaction (chosen randomly from a list of objects), 3) the probability that two hands are visible from the camera at the same time, and 4) the hand that will interact with the object (right or left).

Moreover, we used a Unity asset called \textit{Auto Hand - VR Physics Interaction}\footnote{\url{https://assetstore.unity.com/packages/tools/game-toolkits/auto-hand-vr-physics-interaction-165323}} to improve the physics of the agent when it interacts with the objects. This asset provides a Virtual Reality (VR) interaction system that automatically determines an appropriate hand pose during object manipulation. We have integrated this system into our virtual agent by extending it to automate the grabbing process and adding special types of interactions, such as pressing buttons. Examples of the generated images and poses are reported in Figure~\ref{fig:egoism_synth_anns}.

\section{EgoISM-HOI dataset}\label{sec:egoism_hoi}
We present a new multimodal dataset of EHOIs in the aforementioned industrial scenario called \textit{EgoISM-HOI}. It is composed of two parts: 1) a generated synthetic set of images, and 2) a real-world set of data. Henceforth, we will refer to the synthetic set as \textit{EgoISM-HOI-Synth}, whereas we refer to the real-world data as \textit{EgoISM-HOI-Real}.

\paragraph{\textbf{EgoISM-HOI-Synth}} We adopted our EHOI generation pipeline to generate \textit{EgoISM-HOI-Synth}. It contains a total of 23,356 images with associated depth maps and instance segmentation masks, 35,672 hand instances of which 18,884 are involved in an interaction, and 148,024 object instances across the 19 object categories reported in Figure~\ref{fig:object_models}. Examples of the data which composes the dataset are reported in Figure~\ref{fig:egoism_synth_anns}, while Table~\ref{tab:statistics_egoism} reports statistics about the dataset, including the total number of images, hands, objects, and EHOIs.
\begin{table}[t!]
    \caption{Statistics of \textit{EgoISM-HOI-Synth}.}
    \centering
    \resizebox{\linewidth}{!}{
    \begin{tabular}{lcccccc}
    \hline
    Set     & \#images  & \#hands   & \#EHOIs   & \#left hands  & \#right hands & \#objects \\ \hline
    Train   & 20,788    & 31,790    & 16,786    & 16,019        & 15,771        & 131,968   \\
    Val     & 2,568     & 3,912     & 2,098     & 1,989         & 1,923         & 16,056    \\
    Total   & 23,356    & 35,672    & 18,884    & 18,008        & 17,694        & 148,024   \\ \hline
    \end{tabular}}
    \label{tab:statistics_egoism}
\end{table}

\begin{figure}[t]
    \centering
    \includegraphics[scale=1]{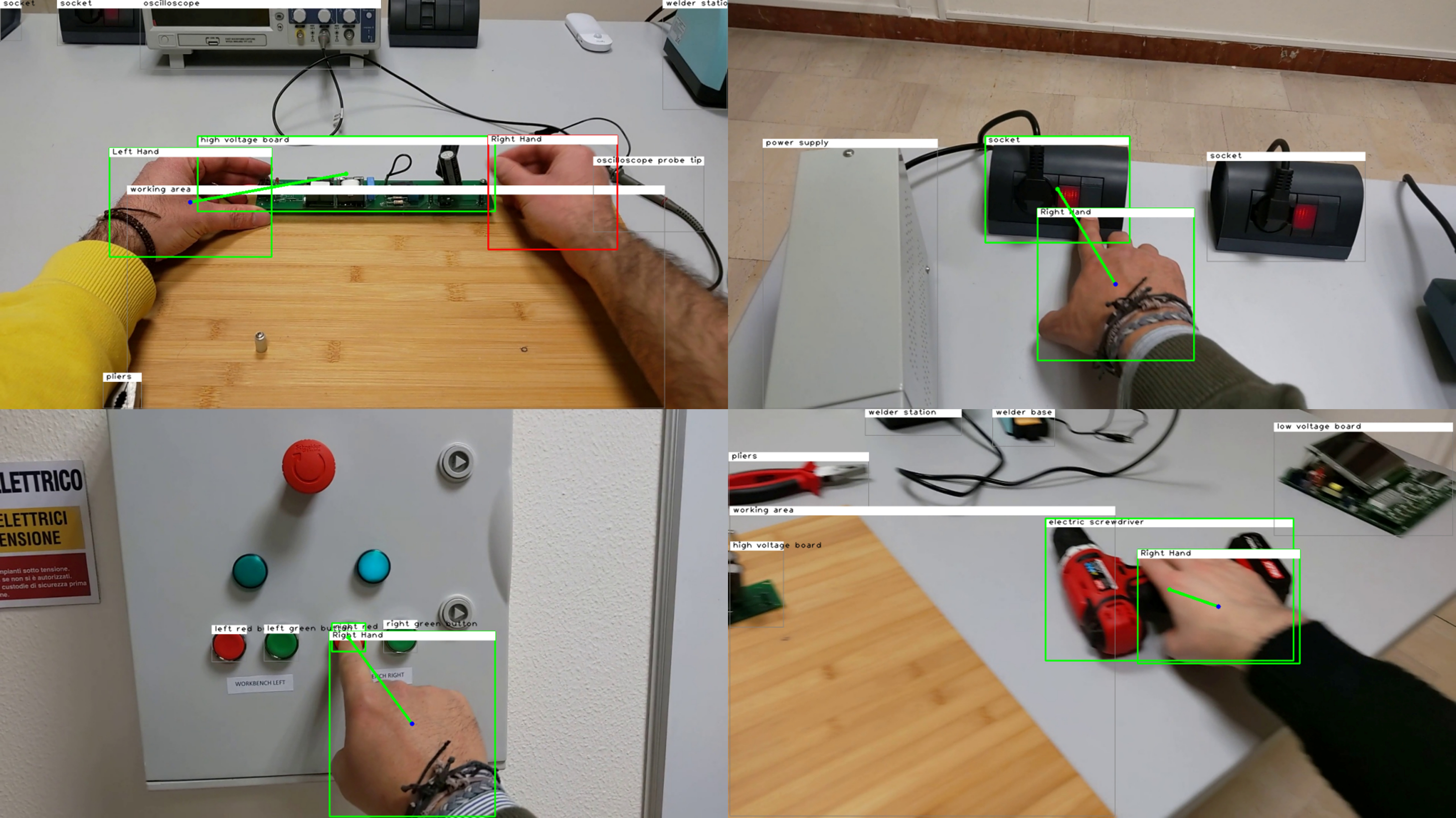}
    \caption{Examples of \textit{EgoISM-HOI-Real} images with the corresponding EHOI annotations.}
    \label{fig:egoism_real_anns}
\end{figure}

\paragraph{\textbf{EgoISM-HOI-Real}} For \textit{EgoISM-HOI-Real}, we expand the previous original real-world dataset \citep{leonardi2022egocentric} from 8 to 42 real egocentric videos in the ENIGMA Laboratory. In these videos, subjects performed testing and repairing operations on electrical boards using laboratory tools. We developed an application for Microsoft Hololens 2, designed to assist operators through the data acquisition process. This application offers audio guidance and shows images to facilitate complex operations during the acquisition phase. Additionally, it integrates voice commands to enhance human-device interaction, allowing operators to give commands such as "next" or "back" to navigate through procedure steps. We defined 8 procedures composed of several steps, in which we vary the tools and electrical boards interacted by the users. Nineteen subjects participated in the data collection. Two were women and seventeen were men. For privacy reasons, we made sure that no other people were visible in the videos, and all the subjects removed any personal object that might reveal their identities (e.g., rings or wristwatches). We acquired 18 hours, 48 minutes, and 13 seconds of video recordings, with an average duration of 26 minutes and 51 seconds, at a resolution of 2272x1278 pixels and a framerate of 30fps. Table~\ref{tab:statistics_real_data} summarizes statistics about the collected data.
\begin{table*}[!t]
    \caption{Statistics of \textit{EgoISM-HOI-Real} data. Since we mainly want to use synthetic data to train models, we used most of the real-world data for testing.}
    \centering
    \resizebox{\linewidth}{!}{
    \begin{tabular}{lcccccccccc}
    \hline
    Set     & \#videos  & \#subjects    &\#procedures   & cumulative videos length  & \#images & \#hands   & \#EHOIs   & \#left hands  & \#right hands & \#objects  \\ \hline
    Train   & 2         & 1             & 2             & 1h:00m:52s                & 1,010    & 1,686     & 1,262     & 758           & 928           & 6,689      \\
    Val     & 10        & 7             & 6             & 4h:35m:28s                & 3,717    & 5,622     & 3,867     & 2,577         & 3,045         & 20,916     \\
    Test    & 30        & 15            & 8             & 13h:11m:51s               & 11,221   & 16,850    & 11,403    & 7,743         & 9,107         & 62,356     \\
    Total   & 42        & 19            & 8             & 18h:48m:13s               & 15,948   & 24,158    & 16,532    & 11,078        & 13,080        & 89,961     \\ \hline
    \end{tabular}}
    \label{tab:statistics_real_data}
\end{table*}
From these videos, we manually annotated 15,948 images following this strategy: 1) we annotated the first frame in which the hand touches an object (i.e., contact frame), and 2) we annotated the first frame after the hand released the object (i.e., end of contact frame). Finally, we assigned the following attributes: 1) hands and objects bounding boxes, 2) hand side (Left/Right), 3) hand contact state (Contact/No contact), 4) hand-object relationships (e.g., hand \textit{x} touches object \textit{y}), and 5) object categories. Figure~\ref{fig:egoism_real_anns} shows some images from this set of data along with the related annotations.
\begin{figure*}[t!]
    \centering
    \includegraphics[scale=1]{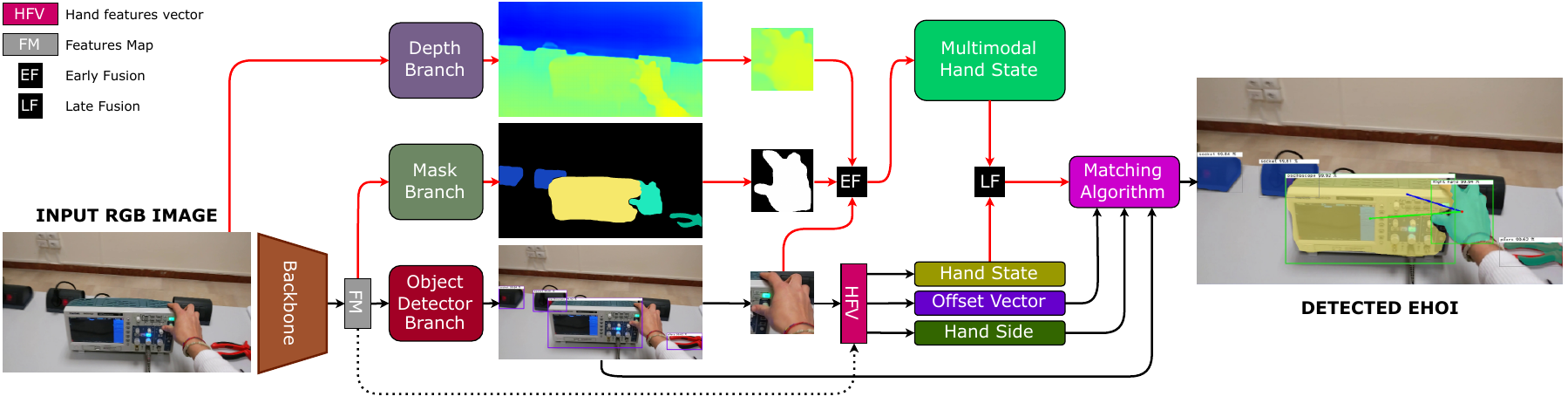}
    \caption{Overall architecture of the proposed Multimodal EHOI detection system. First, the \textit{backbone} extracts image features from the input RGB image. Then, the \textit{object detector branch} and the \textit{instance segmentation branch} detect and generate segmentation masks for all hands and objects in the image. At the same time, the \textit{monocular depth estimation branch} predicts a depth map of the scene. Next, the hand feature vectors obtained through \textit{RoI Pooling} are sent to the following modules for predicting hand attributes: 1) the \textit{hand side classifier}, 2) \textit{hand state classifier}, and 3) \textit{offset vector regressor}. Simultaneously, the RGB image, depth map, and instance segmentation mask of each hand are combined and passed to the \textit{multimodal hand state classifier} module to predict the hand contact state. Finally, the outputs from the previous components are combined and passed to a \textit{matching algorithm} to predict EHOIs.}
    \label{fig:network_ehoi}
\end{figure*}


\section{Proposed approach}\label{sec:approach}
Inspired by \citet{Shan2020UnderstandingHH}, our system extends a two-stage object detector with additional modules specialized to recognize human-object interactions. Differently from our previous work \citep{leonardi2022egocentric}, the proposed method is able to exploit different data signals, such as instance segmentation maps and depth maps, to improve the performance of classic HOI detection approaches. Moreover, our method is able to recognize the class of all the objects in the scene. We believe that this knowledge could be used for other downstream tasks.

Figure~\ref{fig:network_ehoi} shows a diagram of the overall architecture of the method. Firstly, the input RGB image is passed to the \textit{backbone} component to extract the image features. These features are used by the \textit{object detector branch} and the \textit{instance segmentation branch} to detect, recognize and generate segmentation masks of all the objects and hands in the image. Simultaneously, the \textit{monocular depth estimation branch} predicts a depth map of the scene from the RGB image. Then, using the hand boxes predicted by the \textit{object detector branch} and the features map produced by the backbone, the hand feature vectors are extracted with \textit{RoI pooling} and sent to the following modules: 1) the \textit{hand side classifier}, 2) \textit{hand state classifier}, and 3) \textit{offset vector regressor}. These modules predict several hand attributes that will be detailed later. Furthermore, the RGB image, the depth map, and the instance segmentation mask of each hand are combined using an early fusion strategy and passed to the \textit{multimodal hand state classifier} component to predict the hand contact state. As the last step, the resulting outputs of the previous modules are combined and passed to a \textit{matching algorithm} to predict EHOIs in the form of \textit{\textless hand, contact state, active object\textgreater} triplets. The various modules composing our system are described in detail in the following.
\begin{figure*}[t]
    \centering
    \includegraphics[scale=1]{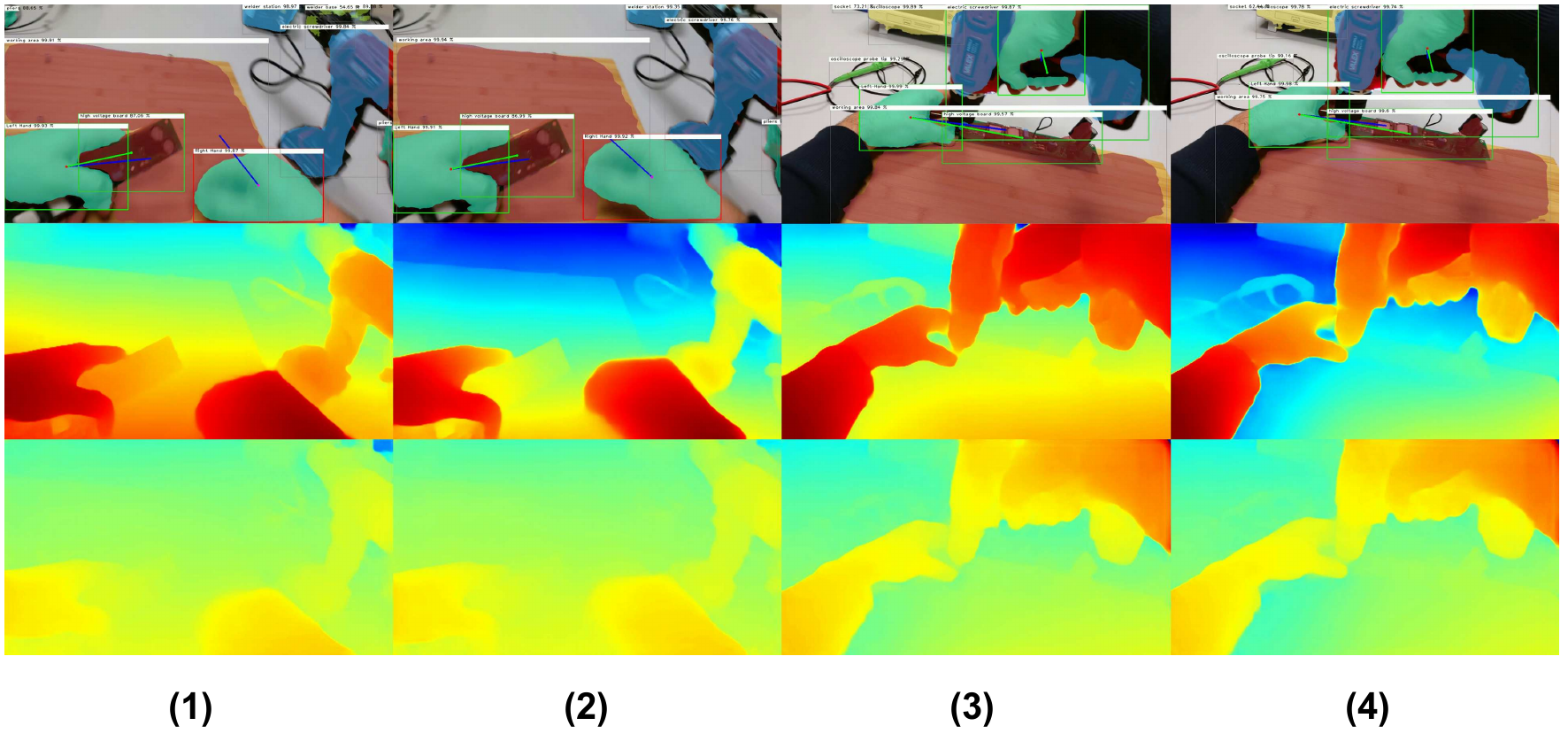}
    \caption{Comparison of the depth maps predicted by our \textit{monocular depth estimation branch}. The first row shows RGB video frames, while the second and third rows contain depth maps predicted by two different models fine-tuned, respectively, by using the losses described in \citet{Ranftl2022} and the proposed one in Equation~(\ref{eq_loss_depth}). The results of the third row are more uniform, while the predicted depth values of the second row vary considerably between similar frames (e.g., the background of (3) and (4) or the object in contact with the left-hand of (1) and (2)).}
    \label{fig:depth_comparison}
\end{figure*}

\paragraph{\textbf{Backbone}} This component consists of a ResNet-101 backbone \citep{he2016deep} with a Feature Pyramid Network (FPN) \citep{lin2017feature}. It takes an RGB image as input and returns a feature map.
\paragraph{\textbf{Object detector branch}} We used Faster-RCNN \citep{faster_rccn}\footnote{We used the following implementation: \url{https://github.com/facebookresearch/detectron2}}, which uses two branches that take as input the features extracted by a backbone to detect and recognize objects and hands in the image.
\paragraph{\textbf{Instance segmentation branch}} We followed Mask-RCNN \citep{mask_rcnn} and add a branch to predict instance segmentation masks from the features extracted by a backbone.
\paragraph{\textbf{Monocular depth estimation branch}} We used the system presented in \citep{Ranftl2022}, called \textit{MiDaS}, to build the monocular depth estimation branch. Given a single RGB image as input, this component estimates the 3D distance to the camera of each pixel. To make the prediction scale of the depth values uniform in our domain, we fine-tuned \textit{MiDaS}\footnote{We used the model \textit{midas\_v21\_384} available in the following repository: \url{https://github.com/isl-org/MiDaS}} redefining the loss function as follows:
\begin{equation}
\label{eq_loss_depth}
\mathcal{L}_{depth}(d,d^*)= \alpha\mathcal{L}_{ssim}(e,e^*) + \beta\mathcal{L}_{ssim}(d,d^*) + \gamma\mathcal{L}_{l1}(d,d^*)
\end{equation}
where $d, d^*$ are the prediction and ground truth depth maps, and $e, e^*$ represent the edge maps of $d,d^*$. $\mathcal{L}_{ssim}$ denotes the \textit{SSIM loss function}, which is used to learn the structure of the depth map, and $\mathcal{L}_{l1}$ is the standard \textit{L1 Loss function} used to learn the depth values of each pixel. Finally, the factors $\alpha$, $\beta$, and $\gamma$ are used to regulate the scale of the $\mathcal{L}_{depth}$ components. During our experiments, we set these factors as follows: $\alpha=0.85$, $\beta=0.9$, and $\gamma=0.9$. 

Differently from the loss proposed in \citep{Ranftl2022}, which standardizes the scale of the depth maps for various datasets, the loss in~\ref{eq_loss_depth} allows the prediction of values convertible into a real 3d distance. Some examples of the considered depth maps are reported in Figure~\ref{fig:depth_comparison}. 

\begin{table*}[t]
	\centering
	\caption{Results of the proposed approach on \textit{EgoISM-HOI-Real} test data. The \textit{EgoISM-HOI-Synth} column indicates whether the \textit{EgoISM-HOI-Synth} training set was used for pre-training models. The \textit{EgoISM-HOI-Real \%} column shows the percentage of real-world data used for fine-tuning. The \textit{improvement} rows show the improvements of models pre-trained with synthetic data compared to models using only real data.}\label{tab:ehoi_results}
	\resizebox{\linewidth}{!}{
		\begin{tabular}{lcccccc} 
			\hline
			\textbf{EgoISM-HOI-Synth} & \textbf{EgoISM-HOI-Real\%} & \textbf{AP Hand}  & \textbf{AP H.+Side} & \textbf{AP H.+State} & \textbf{mAP H.+Obj} & \textbf{mAP H.+All} \\ \hline
			Yes                       & 0                          & 90.02             & 84.72               & 31.85                & 23.92               & 23.28               \\ \hline
			No                        & 10                         & 90.08             & 88.57               & 45.69                & 18.19               & 17.48               \\
			Yes                       & 10                         & 90.53             & 89.34               & 46.64                & 30.90               & 30.65               \\ \hline
												
			\rowcolor{gray!10} \multicolumn{2}{l}{\textit{Improvement over 10\% EgoISM-HOI-Real only}}   & $\textcolor{blue}{\textbf{+0.45}}$ &  $\textcolor{blue}{\textbf{+0.77}}$   & $\textcolor{blue}{\textbf{+0.95}}$  & $\textcolor{blue}{\textbf{+12.71}}$ & $\textcolor{blue}{\textbf{+13.17}}$ \\ \hline						
												
			No                        & 25                         & 90.43             & 89.45               & 43.73                & 18.72               & 18.31               \\
			Yes                       & 25                         & 90.66             & 89.71               & 48.31                & 31.76               & 31.33               \\ \hline
			\rowcolor{gray!10} \multicolumn{2}{l}{\textit{Improvement over 25\% EgoISM-HOI-Real only}}  & $\textcolor{blue}{\textbf{+0.23}}$ &  $\textcolor{blue}{\textbf{+0.26}}$   & $\textcolor{blue}{\textbf{+4.58}}$  & $\textcolor{blue}{\textbf{+13.04}}$ & $\textcolor{blue}{\textbf{+13.02}}$ \\ \hline
												
			No                        & 50                         & 90.43             & 89.57               & 52.74                & 19.17               & 19.06               \\
			Yes                       & 50                         & \underline{90.69} & \underline{90.00}   & 54.79                & \underline{34.12}   & \underline{33.12}   \\ \hline
			\rowcolor{gray!10} \multicolumn{2}{l}{\textit{Improvement over 50\% EgoISM-HOI-Real only}}   & $\textcolor{blue}{\textbf{+0.26}}$ &  $\textcolor{blue}{\textbf{+0.43}}$   & $\textcolor{blue}{\textbf{+2.05}}$  & $\textcolor{blue}{\textbf{+14.95}}$ & $\textcolor{blue}{\textbf{+14.06}}$ \\ \hline
			
			No                        & 100                        & 90.54             & \textbf{90.06}      & \underline{56.34}    & 22.31               & 21.76               \\ 
			Yes                       & 100                        & \textbf{90.73}    & 89.99               & \textbf{56.88}       & \textbf{35.94}      & \textbf{35.47}      \\ \hline
			\rowcolor{gray!10} \multicolumn{2}{l}{\textit{Improvement over 100\% EgoISM-HOI-Real only}}   & $\textcolor{blue}{\textbf{+0.19}}$ &  $\textcolor{red}{-0.07}$   & $\textcolor{blue}{\textbf{+0.54}}$  & $\textcolor{blue}{\textbf{+13.63}}$ & $\textcolor{blue}{\textbf{+13.71}}$ \\ \hline
			         
		\end{tabular}}
\end{table*}

\paragraph{\textbf{Hand side classifier}} A Multi-Layer Perceptron (MLP) with a hidden fully connected layer that takes as input an ROI-pooled feature vector of the hand crop to predict the hand side (\textit{left/right}).
\paragraph{\textbf{Hand state classifier}} This module classifies the contact state of the detected hands through an additional MLP with a hidden fully connected layer. It takes as input the hand features vector, enlarged by 30\% to include information about the surrounding context (e.g., nearby objects), and predicts the hand contact state (\textit{no contact/in contact}).
\paragraph{\textbf{Multimodal hand state classifier}} This component is based on the EfficientNetV2 architecture \citep{TanL21}. It takes as input a combination of RGB, depth map (inferred by the \textit{monocular depth estimation branch}), and instance segmentation mask (predicted by the \textit{instance segmentation branch}) of each hand to estimate the hand contact state. The output of this module is combined with the output of the \textit{hand state classifier} to obtain the final prediction of the hand contact state.
\paragraph{\textbf{Offset vector regressor}} This module infers a vector that links the center of the bounding box of each hand to the center of the bounding box of the candidate active object (i.e., the object touched by the hand). This module consists of an MLP which takes as input the ROI-pooled feature vectors of the hands to predict \textit{\textless $v_x$, $v_y$, m\textgreater} triplets, where ($v_x, v_y$) represent the direction of the vector and $m$ its magnitude.
\paragraph{\textbf{Matching algorithm}} The final module of our system is a matching algorithm that exploits the outputs of the previous modules to predict EHOIs as \textit{\textless hand, contact state, active object\textgreater} triplets. For each detected hand, the algorithm calculates an interaction point ($p_{ehoi}$) using the bounding box center of the hand and the corresponding offset vector. $p_{ehoi}$ represents the prediction of the bounding box center of the candidate active object. Finally, the object whose center is closest to $p_{ehoi}$ is chosen as the active object. \\

To optimize our system during the training phase, we used the standard Faster R-CNN loss \citep{faster_rccn} for the \textit{object detector branch}, while we utilized the definition of \citep{mask_rcnn} for the \textit{instance segmentation branch}. As previously discussed, to optimize the \textit{monocular depth estimation branch} we exploited the loss function in Equation~(\ref{eq_loss_depth}). We used the standard \textit{binary cross-entropy loss} for the \textit{hand side classifier}, whereas for \textit{offset vector regressor} we used the \textit{mean squared error loss}. We optimized the \textit{hand state classifier} and \textit{multimodal hand state classifier} according to the following equation:
\begin{equation}
\label{eq_loss_cs}
\resizebox{0.95\linewidth}{!}{$
\mathcal{L}_{cs}(cs, cs^*)= \mathcal{L}_{bce}(cs_{rgb}, cs^*) + \mathcal{L}_{bce}(cs_{mm}, cs^*) + \mathcal{L}_{bce}(cs_{lf}, cs^*)
$}
\end{equation}
where $cs, cs^*$ are the prediction and ground truth hand contact states, $cs_{rgb}$, $cs_{mm}$, and $cs_{lf}$ denotes, respectively, the predictions of the hand contact states of the \textit{hand state classifier}, \textit{multimodal hand state classifier} and the combined predictions of these modules. $\mathcal{L}_{bce}$ denotes the standard \textit{binary cross-entropy loss}. The final loss of our system is the sum of all the aforementioned losses.

\section{Experimental results}\label{sec:experimental_results}
We conducted a series of experiments to 
1) assess how much the generated synthetic data are useful in training models able to generalize to the real-world domain (Section~\ref{sec:ehoi_results}),
2) highlight the contribution of multimodal signals to tackle the EHOI detection task (Section~\ref{sec:multimodal_ehoi_results}), and
3) compare the proposed method with a set of baselines based on state-of-the-art class-agnostic approaches (Section~\ref{sec:comparison_sota}). Section~\ref{sec:additional_results} reports ablation studies, delving into the contributions of various modules and the impact of different volumes of synthetic images on our model's performance. Moreover, it reports additional results on pre-training our method with external data and improvements obtained by our approach for the object detection task.

\subsection{Experimental Settings}
\paragraph{\textbf{Dataset}} We performed experiments on the proposed \textit{EgoISM-HOI} dataset. Since we want to exploit synthetic data to train models to detect EHOIs when few or zero real-world data are available, we used the splits reported in Table~\ref{tab:statistics_egoism} and Table~\ref{tab:statistics_real_data} for the synthetic and real data respectively.
\paragraph{\textbf{Evaluation Metrics}} Following \citet{Shan2020UnderstandingHH}, we evaluated our method using metrics based on standard \textit{Average Precision}, which assess the models' ability to detect hands and objects as well as the correctness of some attributes such as the hand state, the hand side, and whether an object is active (i.e., it is involved in an interaction). In addition, since our model predicts active object classes, we computed the \textit{mean Average Precision} (mAP) to consider the correctness of the predicted object classes. Specifically, we used the following metrics: 1) \textit{AP Hand}: \textit{Average Precision} of the hand detections, 2) \textit{AP Hand+Side}: \textit{Average Precision} of the hand detections considering the correctness of the hand side, 3) \textit{AP Hand+State}: \textit{Average Precision} of the hand detections considering the correctness of the hand state, 4) \textit{mAP Hand+Obj}: \textit{mean Average Precision} of the \textit{\textless hand, active object\textgreater} detected pairs, and 5) \textit{mAP Hand+All}: combinations of  \textit{AP Hand+Side}, \textit{AP Hand+State}, and \textit{mAP Hand+Obj} metrics.
\paragraph{\textbf{Training Details}} To perform all the experiments we used a machine with a single \textit{NVIDIA A30} GPU and an \textit{Intel Xeon Silver 4310} CPU. We scaled images for both the training and inference phases to a resolution of 1280x720 pixels. We trained models on \textit{EgoISM-HOI-Synth} with \textit{Stochastic Gradient Descent} (SGD) for 80,000 iterations with an initial learning rate equal to 0.001, which is decreased by a factor of 10 after 40,000 and 60,000 iterations, and a minibatch size of 4 images. Instead, to fine-tune the models with \textit{EgoISM-HOI-Real} training data, we froze the \textit{monocular depth estimation branch} and \textit{instance segmentation branch} modules. Finally, we trained the models for 20,000 iterations and decreased the initial learning rate (0.001) by a factor of 10 after 12,500 and 15,000 iterations.

\subsection{The Impact of Synthetic Data on System Performance}\label{sec:ehoi_results}
The goal of this set of experiments is to show the ability of a model trained with synthetic data to generalize to real-world data. Specifically, we want to demonstrate how the synthetic data generated by the proposed tool can be used to represent realistic human-object interactions.

We compared models pre-trained on the \textit{EgoISM-HOI-Synth} training split and fine-tuned using different amounts of \textit{EgoISM-HOI-Real} training data (i.e., 0\%, 10\%, 25\%, 50\%, and 100\%) with models trained only with \textit{EgoISM-HOI-Real} data. Since the \textit{multimodal hand state classifier}, \textit{monocular depth estimation branch}, and \textit{instance segmentation branch} modules need to be trained with labels available only on synthetic data, we deactivated these components in all the models in this set of experiments for a fair comparison.

\begin{figure}[t]
    \centering
    \includegraphics[scale=1]{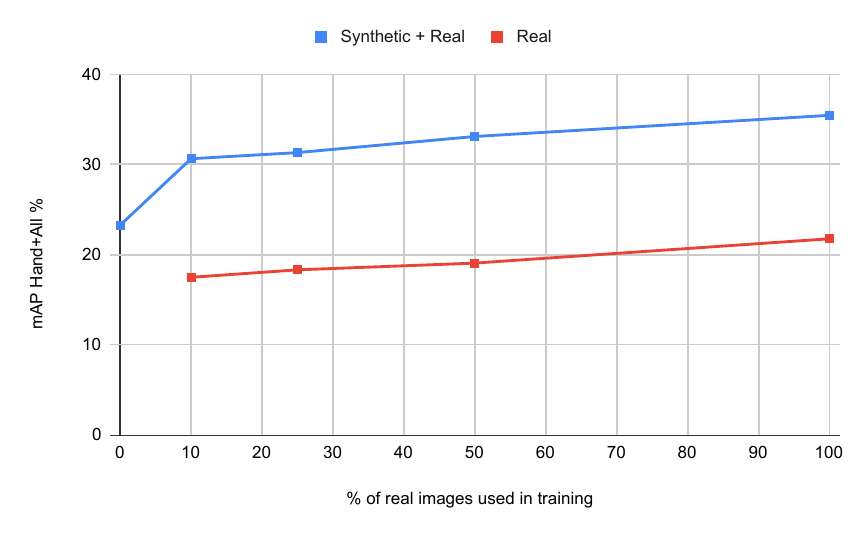}
    \caption{Performance comparison of the proposed system on our \textit{EgoISM-HOI-Real} test data in terms of \textit{mAP Hand+All}. The blue curve reports the results of the models pre-trained on \textit{EgoISM-HOI-Synth} and fine-tuned at different percentages of the \textit{EgoISM-HOI-Real} training set, while the red curve reports the results of the models trained on real-world data only.}
    \label{fig:graphical_ehoi_results}
\end{figure}

Table~\ref{tab:ehoi_results} reports EHOI detection results on the \textit{EgoISM-HOI-Real} test set. Models pre-trained with \textit{EgoISM-HOI-Synth} data outperform all the corresponding models trained using only \textit{EgoISM-HOI-Real} data by consistent margins according to all the evaluation metrics. Considering the two models fine-tuned using the 100\% of the \textit{EgoISM-HOI-Real} training set, the improvements of the model pre-trained with \textit{EgoISM-HOI-Synth} data are significant in the evaluation measures affected by active objects (i.e., \textit{mAP Hand+Obj} and \textit{mAP Hand+All}). Specifically, there is an increase of +13.63\% (35.94\% vs. 22.31\%) for \textit{mAP Hand+Obj} and of +13.71\% (35.47\% vs. 21.76\%) for \textit{mAP Hand+All}. The improvements persist across all the other configurations of real-world training data, i.e., 10\%, 25\%, and 50\%. Models pre-trained with synthetic data show considerable increments of performances of +13.17\%, +13.02\%, and +14.06\%, respectively, in terms of \textit{mAP Hand+All}, compared to models trained only on \textit{EgoISM-HOI-Real} data.

Considering the \textit{AP Hand}, \textit{AP H.+Side} and \textit{AP Hand+State} evaluation measures, we observe marginal enhancements in the performance of models pre-trained on \textit{EgoISM-HOI-Synth}. These results suggest that using synthetic data for pre-training models significantly improves the method's capability to detect active objects, which are susceptible to frequent occlusions by the hands. Measures influenced only by hands show minimal benefit from additional real-world data, suggesting that they reach a saturation point earlier in terms of performance improvement.

In addition, it is worth noting that the model trained using only the \textit{EgoISM-HOI-Synth} data (row 1) outperforms the best model that used only the real-world data for the evaluation measures influenced by the active objects, obtaining +1,61\% (23.92\% vs 22.31\%) and +1,52\% (23.28\% vs 21.76\%) for the \textit{mAP Hand+Obj} and \textit{mAP Hand+All} measures respectively. Figure~\ref{fig:graphical_ehoi_results} further illustrates the results in terms of \textit{mAP Hand+All} considering different amounts of \textit{EgoISM-HOI-Real} training data in the fine-tuning.

\subsection{Impact of Multimodal training}\label{sec:multimodal_ehoi_results} 
This set of experiments aims to highlight the contribution of the different modalities involved in our approach. 
For these experiments, we consider the full architecture illustrated in Figure~\ref{fig:network_ehoi} comprising the \textit{backbone}, the \textit{object detector branch}, the \textit{instance segmentation branch}, the \textit{monocular depth estimation branch}, and the \textit{multimodal hand state classifier}.
As a baseline, we considered a model trained by deactivating the \textit{multimodal hand state classifier}, \textit{monocular depth estimation branch}, and \textit{instance segmentation branch} modules. We compare this baseline with several versions of the proposed architecture in which the \textit{hand contact state} is estimated using different subsets of modalities (i.e., RGB, Depth, and Mask) and modules (i.e., \textit{multimodal hand state classifier}, and \textit{hand state classifier}). As these modules only affect the prediction of hand contact state, Table~\ref{tab:multimodal_ehoi_results} reports only the metrics affected by these predictions (i.e., \textit{AP Hand+State} and \textit{mAP Hand+All}). Note that all the models used in this experiment were pre-trained using \textit{EgoISM-HOI-Synth} and then fine-tuned using 100\% of the \textit{EgoISM-HOI-Real} training set. 

\begin{table}[t]
    \centering
    \caption{Experiments to evaluate the impact on system performance of the different modalities and components involved in our architecture. The \textit{Contact state} column indicates the branches used to predict the \textit{hand contact states}, i.e., \textit{multimodal hand state classifier} (MHS), and \textit{Hand state classifier} (HS). While the \textit{MHS Input Modalities} column indicates the modalities passed in input to the \textit{multimodal hand state classifier}. The best results are highlighted in bold, whereas the second-best results are underlined.}\label{tab:multimodal_ehoi_results}
    \resizebox{\linewidth}{!}{
    \begin{tabular}{lcccc}
        \hline
        \textbf{Contact state}               & \textbf{MHS Input Modalities}                  & \textbf{AP H+State}        & \textbf{mAP H+All} \\ \hline
        HS                          & -                                     & 56.88             & 35.47 \\ 
        HS+MHS                      & RGB                                   & 58.29             & 35.71 \\
        HS+MHS                      & RGB+DEPTH                             & \underline{58.37} & \underline{35.92} \\
        HS+MHS                      & RGB+MASK                              & 58.30             & 35.34 \\
        HS+MHS                      & RGB+DEPTH+MASK                        & \textbf{58.40}    & \textbf{36.51} \\
        MHS                         & RGB+DEPTH+MASK                        & 57.56             & 35.81 \\ \hline
    \end{tabular}}
\end{table}

\begin{table}[t]
    \centering
    \caption{Comparison between the proposed system and different baseline approaches based on HIC \citep{Shan2020UnderstandingHH}.}
    \label{tab:comparison_hic}
    \resizebox{\linewidth}{!}{
    \begin{tabular}{lccccccc}
    \hline
    \textbf{Method}              & \textbf{EgoISM-HOI-Synth} & \textbf{EgoISM-HOI-Real\%}    & \textbf{mAP Hand+All} \\ \hline  
    Proposed (Base)     & Yes                       & 0                     & 23.28 \\
    Proposed (Base)     & Yes                       & 10                    & \underline{30.65} \\
    Proposed (Full)     & Yes                       & 100                   & \textbf{36.51} \\    
    HIC+RESNET (BS1)    & No                        & 100*                  & 09.92 \\             
    HIC+RESNET (BS2)    & No                        & 100                   & 22.18 \\             
    HIC+RESNET (BS3)    & Yes                       & 0                     & 16.39  \\            
    HIC+RESNET (BS4)    & Yes                       & 100                   & 23.59 \\ 
    HIC+YOLOv5 (BS5)    & Yes                       & 100                   & 20.62 \\ \hline     
    \end{tabular}}
\end{table}

\begin{figure*}[t]
    \centering
    \includegraphics[]{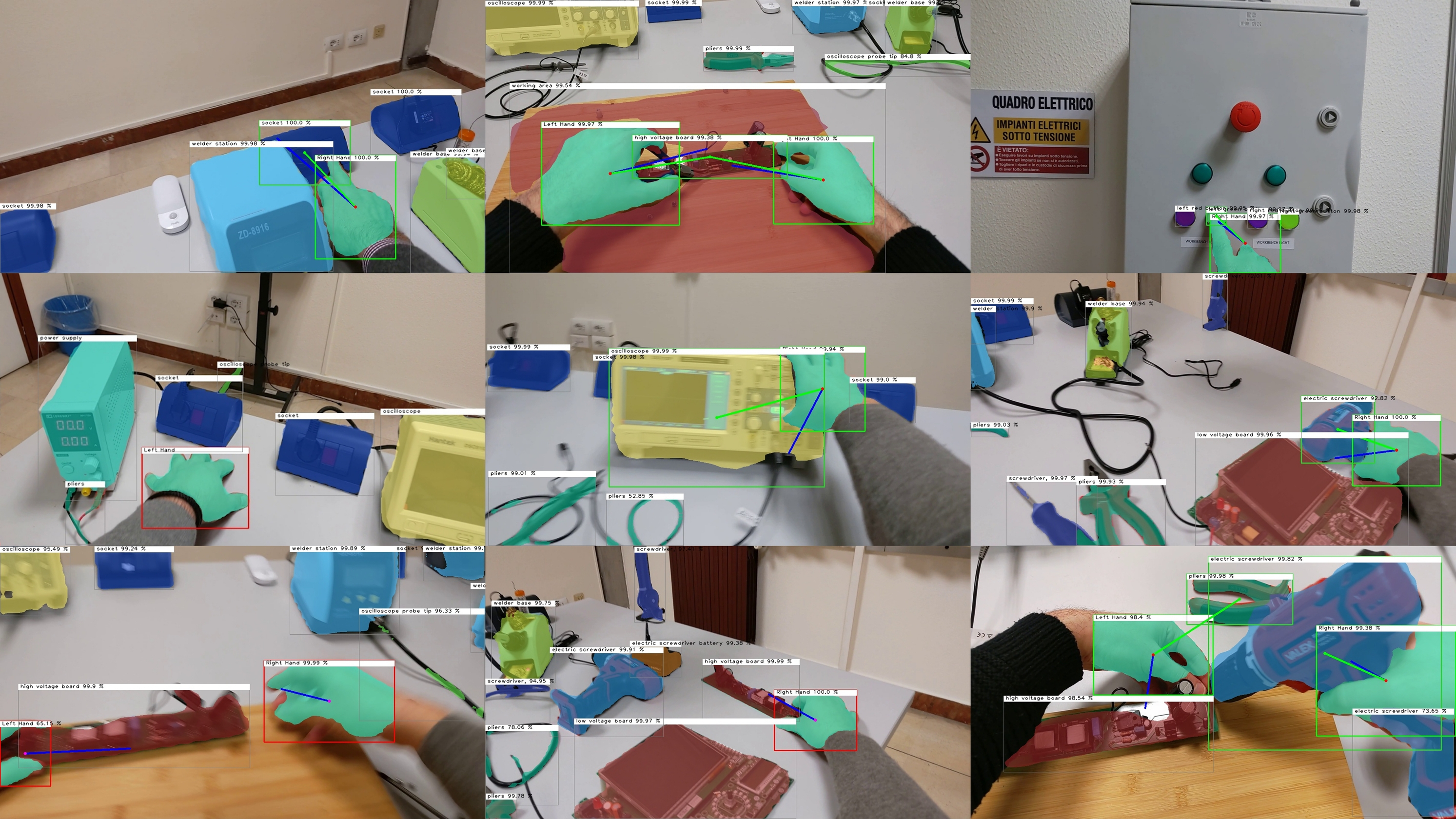}
    \caption{Qualitative results of the proposed multimodal EHOI detection system on the \textit{EgoISM-HOI-Real} test data.}
    \label{fig:qualitative_examples}
\end{figure*}

Combining the predictions of the \textit{multimodal hand state classifier} and \textit{hand state classifier} modules (rows 2-5) leads to general improvements in the system performance over the models that use only a single branch to predict the \textit{hand contact state} (rows 1 and 6), with maximum improvements over the baseline (rows 5 vs 1) of +1,52\% (58.40 vs 56.88) for the \textit{AP Hand+State} and +1.04\% (36.51 vs 35.47) for the \textit{mAP Hand+All}. Fusing RGB with Depth signals (row 3) brings a small improvement of +0.21\% (35.92 vs 35.71) for the \textit{mAP Hand+All} over the model which uses only the RGB signal (row 2). Interestingly, combining RGB with Mask (row 4) improves the result of +1.42\% (58.30 vs 56.88) over the baseline (row 1) in terms of \textit{AP Hand+State} but leads to a worsening performance of -0.13\% (35.34 vs 35.47) considering the \textit{mAP Hand+All} measure. This suggests that the method is unable to benefit from segmentation masks in the absence of the depth signal. Finally, fusing all the modalities (row 5) leads to the best performance, bringing an improvement over the second-best result (RGB+DEPTH, row 3) of +0.59\% (36.51 vs 35.92) for the \textit{mAP Hand+All} metric. Figure~\ref{fig:qualitative_examples} shows some qualitative results obtained with the full proposed architecture.

\subsection{Comparison with class-agnostic baselines}\label{sec:comparison_sota}
This section compares our proposed approach with different baseline approaches based on state-of-the-art methods \citep{Shan2020UnderstandingHH, VISOR2022}. Specifically, we compare our approach with the method of \citet{Shan2020UnderstandingHH} in section~\ref{sec:comparison_hic} and with the method of \citet{VISOR2022} in section~\ref{sec:comparison_visor}.

\subsubsection{Comparison with HiC}\label{sec:comparison_hic}

Table~\ref{tab:comparison_hic} compares our system with different instances of the class-agnostic method introduced in \citet{Shan2020UnderstandingHH}. Henceforth, we will refer to this method as \textit{Hands In Contact} (HIC). Since HIC is class agnostic, to compare our method with it, we extend it to recognize the active object classes following two different approaches. In the first approach, we used a Resnet-18 CNN \citep{he2015deep} to classify image patches extracted from the active object bounding boxes. We trained the classifier with four different sets of data: 1) \textit{BS1}: we sampled 20,000 frames from 19 videos where a single object of each class is shot at a time. This collection provides a minimal training set that can be collected with a modest labeling effort (comparable with the time needed for acquiring 3D models of the objects in our pipeline); 2) \textit{BS2}: we used images from the proposed \textit{EgoISM-HOI-Real} training set; 3) \textit{BS3}: we used images from the proposed \textit{EgoISM-HOI-Synth} training set; 4) \textit{BS4}: we used all \textit{EgoISM-HOI} data. The second approach (\textit{BS5}) exploits a YOLOv5\footnote{YOLOv5: \url{https://github.com/ultralytics/yolov5}} object detector, trained to recognize the considered objects (see Fig.~\ref{fig:object_models}), to assign a label to the active objects predicted by HIC. Specifically, for each active object prediction, we select the class of the object with the highest \textit{IoU} among those predicted by the YOLOv5 object detector or discard the proposal if there are no box intersections. It is worth noting that HIC was pre-trained on the large-scale dataset \textit{100DOH}, which contains over 100K labeled frames of HOIs.

The best model of the proposed EHOI detection method (row 3) outperforms all the baselines (rows 4-8) with significant improvements ranging from +12.92\% (36.51 vs 23.59) to +26,59\% (36.51 vs 9.92). The approach based on Resnet-18 (rows 4-7) leads to better performance compared to the method based on the YOLOv5 object detector (row 8). Indeed, considering only the baselines (rows 4-8), the best result is achieved by BS4 (row 7), which was pre-trained using synthetic and real-world \textit{EgoISM-HOI} data, with an improvement of +2.97\% (23.59 vs 20.62) over the BS5 (row 8). Interestingly, even the BS2 (row 5), which did not use synthetic data during training, obtained a higher result of +1.56\% (22.18 vs 20.62) than the BS5 (row 8). These results suggest the limits of this simple approach. In addition, it is worth noting that the model pre-trained on \textit{EgoISM-HOI-Synth} and fine-tuned using 10\% of the \textit{EgoISM-HOI-Real} training set (row 2) outperforms all the baseline approaches (rows 4-8), with an improvement of +7,06\% (30.65 vs 23.59) over the BS4 (row 7). It is worth mentioning that the model trained only on \textit{EgoISM-HOI-Synth} (row 1) achieves comparable results to the best baseline approach (row 7).

\subsubsection{Comparison with VISOR baseline}\label{sec:comparison_visor}
\begin{table}[t]
	\centering
	\caption{Comparison between the proposed approach and VISOR baseline \citep{Shan2020UnderstandingHH} on \textit{EgoISM-HOI} and \textit{VISOR} datasets.}
	\label{tab:comparison_visor}
	\resizebox{\linewidth}{!}{
		\begin{tabular}{lccccccc}
			\hline
			\textbf{Method} & \textbf{Training set} & \textbf{Test set}    & \textbf{AP Hand+All} \\ \hline 
												    
			VISOR baseline  & EgoISM-HOI-Synth      & EgoISM-HOI-Real test & 10.79                \\ 
			Proposed        & EgoISM-HOI-Synth      & EgoISM-HOI-Real test & 20.47                \\ \hline
			\rowcolor{gray!10} \multicolumn{3}{l}{\textit{Improvement}} & $\textcolor{blue}{\textbf{+9.68}}$ \\ \hline
												    
			VISOR baseline  & EgoISM-HOI-Real       & EgoISM-HOI-Real test & 26.58                \\ 
			Proposed        & EgoISM-HOI-Real       & EgoISM-HOI-Real test & 33.17                \\ \hline
			\rowcolor{gray!10} \multicolumn{3}{l}{\textit{Improvement}} & $\textcolor{blue}{\textbf{+6.59}}$ \\ \hline
												    
			VISOR baseline  & EgoISM-HOI            & EgoISM-HOI-Real test & 27.60                \\ 
			Proposed        & EgoISM-HOI            & EgoISM-HOI-Real test & 32.13                \\ \hline
			\rowcolor{gray!10} \multicolumn{3}{l}{\textit{Improvement}} & $\textcolor{blue}{\textbf{+4.53}}$ \\ \hline 
						
			VISOR baseline  & VISOR                 & VISOR validation     & 44.76                \\ 
			Proposed        & VISOR                 & VISOR validation     & 46.97                \\ \hline
			\rowcolor{gray!10} \multicolumn{3}{l}{\textit{Improvement}} & $\textcolor{blue}{\textbf{+2.21}}$ \\ \hline 
		\end{tabular}}
\end{table}

Table~\ref{tab:comparison_visor} compares our framework with respect to the class-agnostic HOI detection baseline introduced in \citet{VISOR2022}. To comprehensively assess the effectiveness of our proposed approach, we conducted experiments on both our dataset and the validation set of the VISOR dataset. Note that we used the VISOR validation set because, at the time of writing, the test set has not been released publicly. Finally, considering that active object categories weren't labelled in VISOR, to ensure fairness in our comparison, we have employed our approach with an agnostic setting, hence without considering object categories at inference time.
In all configurations, our approach outperformed the VISOR baseline by a significant margin. Specifically, for the test set \textit{EgoISM-HOI-Real} we observed +9.68\%, +6.59\%, +4.53\% when \textit{EgoISM-HOI-Synth}, \textit{EgoISM-HOI-Real}, \textit{EgoISM-HOI} are used as training sets respectively. Furthermore, considering the validation set of the VISOR dataset, our approach surpassed the VISOR baseline by +2.21\%. These results further confirm the effectiveness of the proposed method.

\begin{table*}[t]
	\centering
	\caption{EHOI detection results on \textit{EgoISM-HOI-Real} test data of models pre-trained on 100DOH and VISOR datasets. The grey row shows the results obtained by the model trained only on the EgoISM-HOI dataset, serving as the baseline for this set of experiments.}
	\label{tab:sota_pretraining}
	\resizebox{\linewidth}{!}{
		\begin{tabular}{llccccc}
			\hline
			\textbf{Pre-training}               & \textbf{Fine-tuning} & \textbf{AP Hand} & \textbf{AP H+Side} & \textbf{AP H+State} & \textbf{mAP H+Obj} & \textbf{mAP H+All} \\ \hline
									
			\rowcolor{gray!10} EgoISM-HOI-Synth & EgoISM-HOI-Real      & 90.73            & 89.99              & 56.88               & 35.94              & 35.47              \\ \hline \hline
						   
			100DOH                              & EgoISM-HOI-Synth     & 90.78            & 89.88              & 35.46               & 23.47              & 23.19              \\
			100DOH                              & EgoISM-HOI-Real      & \textbf{90.87}   & 90.44              & \textbf{59.25}      & 18.10              & 17.69              \\
			100DOH                              & EgoISM-HOI           & 90.86            & \textbf{90.51}     & 58.87               & \textbf{38.54}     & \textbf{37.37}     \\ \hline \hline
															        
			VISOR                               & EgoISM-HOI-Synth     & 90.58            & 89.24              & 36.07               & 24.78              & 24.49              \\       
			VISOR                               & EgoISM-HOI-Real      & 90.65            & 90.28              & \textbf{57.84}      & 29.51              & 29.40              \\      
			VISOR                               & EgoISM-HOI           & \textbf{90.74}   & \textbf{90.35}     & 56.71               & \textbf{39.11}     & \textbf{38.95}     \\ \hline    
		\end{tabular}
	}
\end{table*}

\begin{table*}[t]
	\centering
	\caption{Impact of pretraining with varying numbers of \textit{EgoISM-HOI-Synth} images.}\label{tab:numer_of_synthetic}
	\resizebox{\linewidth}{!}{
		\begin{tabular}{lccc} 
			\hline
			\textbf{Pretraining Data} & \textbf{\#Pretraining Images} & \textbf{Finetuning Data} & \textbf{mAP H.+All} \\ \hline
			EgoISM-HOI-Synth & 1,010  & EgoISM-HOI-Real & 32.83 \\ 
			EgoISM-HOI-Synth & 2,020  & EgoISM-HOI-Real & 33.01 \\
			EgoISM-HOI-Synth & 4,040  & EgoISM-HOI-Real & 34.96 \\
			EgoISM-HOI-Synth & 10,100 & EgoISM-HOI-Real & \underline{35.34} \\ 
			EgoISM-HOI-Synth & 20,788 & EgoISM-HOI-Real & \textbf{35.47} \\ \hline
		\end{tabular}}
\end{table*}

\subsection{Additional results}
\label{sec:additional_results}
In this section, we show an additional set of experiments with the aim of 1) demonstrating how using domain-specific synthetic data improves the performance of a system pre-trained on out-of-domain large-scale datasets (Section~\ref{sec:pretraining_datasets}), 2) analyzing the impact of varying quantities of synthetic images on the pre-training of the system (Section~\ref{sec:quantities_synthetic_data}), 3) investigating the influence of the different modules within our system (Section~\ref{sec:weight_modalities}), 4) showing the potential of using synthetic data for the related task of \textit{Object Detection} (Section~\ref{sec:object_detection}). 

\subsubsection{Pre-training on 100 Days Of Hands and VISOR}\label{sec:pretraining_datasets}
To further confirm the usefulness of synthetic data, we performed additional experiments where we pre-trained models on 100DOH \citep{Shan2020UnderstandingHH} and VISOR \citep{VISOR2022} datasets which were then fine-tuned considering the proposed EgoISM-HOI dataset. These experiments aim to demonstrate how leveraging domain-specific synthetic data enhances the performance of a system pre-trained on a large amount of out-of-domain real-world data. The results shown in Table~\ref{tab:sota_pretraining} highlight that employing synthetic data in the fine-tuning phase consistently led to superior performance for both 100DOH and VISOR pre-trained models. Considering the 100DOH pre-trained model, the combination of synthetic data with real-world data (row 4) significantly enhances metrics influenced by active objects (i.e., \textit{mAP Hand+Obj} and \textit{mAP Hand+All}). Specifically, we observed an improvement of +20.44\% (38.54\% vs. 18.10\%) for the \textit{mAP Hand+Obj} and of +19.68\% (37.37\% vs. 17.69\%) for the \textit{mAP Hand+All}, compared to the model trained only on real-world data (row 3). Similar considerations can be made for the experiments performed considering the VISOR dataset as pre-training. Indeed, the model trained using both synthetic data with real-world data (row 7) outperforms their counterparts trained only on real-world data (row 6). In this last case, we observe an improvement of +9.60\% (39.11\% vs. 29.51\%) for \textit{mAP Hand+Obj} and of +9.05\% (38.95\% vs. 29.40\%) for \textit{mAP Hand+All}. It is important to note that, as seen in previous investigations (see Section~\ref{sec:ehoi_results}), for metrics influenced exclusively by hands (for example, \textit{AP Hand}, \textit{AP H+Side} and \textit{AP H+State}), we observed minor or no improvements when synthetic data were used, compared to models trained exclusively on real-world data. Lastly, it's worth noting that the results obtained from the baseline model, i.e. the model trained only on the EgoISM-HOI data (first row), obtained comparable or even superior performance with respect to models trained using real-world data (i.e., 100DOH and VISOR). This further highlights the usefulness of synthetic data in enhancing HOI model performances.

\subsubsection{Effect of Varying Synthetic Data Quantities on Pretraining}\label{sec:quantities_synthetic_data}
We conducted a series of experiments using a varying number of synthetic images from EgoISM-HOI-Synth during pre-training. Table~\ref{tab:numer_of_synthetic} collects the results of this experiment. The results in terms of \textit{mAP Hand+All} show an evident trend of improvement in performances as the number of pre-training images increases. Specifically, starting with 1,010 synthetic images, the model achieved an initial performance of 32.83\%. Subsequently, by doubling the pre-training images to 2,020, a slight improvement was observed, reaching 33.01\%. Performance further increased with 4,040 synthetic images (34.96\%) and with 10,100 synthetic images (35.34\%). Finally, using all the 20,788 synthetic images resulted in the highest performance with a \textit{mAP Hand+All} of 35.47\%. This highlights the impact of a larger synthetic dataset on enhancing the model's performance.

\subsubsection{Understanding the Weighting of Various Modalities in the System}\label{sec:weight_modalities}
Our proposed approach combines the predictions of the \textit{hand contact state} produced by the \textit{hand side classifier} and \textit{multimodal hand state classifier} modules to produce the final \textit{hand contact state} prediction. We have analyzed the impact of these modules in our framework by assigning a weight to the prediction of each module. Table~\ref{tab:weigth_modalities} collect the results of these experiments in terms of \textit{AP Hand+State} and \textit{mAP Hand+All} evaluation measures. The obtained results show that as the weight shifts towards a balance between HS and MHS modules, the performances improve. This highlights the importance of both modules. Specifically, the system achieves its peak performance at 50\% for both modules, obtaining 58.40\% and 36.51\% for the textit{AP Hand+State} and \textit{mAP Hand+All}, respectively.
\begin{table}[t]
	\centering
	\caption{The table demonstrates how varying the weighting between the \textit{hand side classifier} (HS) and the \textit{multimodal hand state classifier} (MHS) affects the performance of the system in terms of \textit{AP Hand+State} and \textit{mAP Hand+All}.}\label{tab:weigth_modalities}
	\resizebox{\linewidth}{!}{
		\begin{tabular}{lccc} 
			\hline
			\textbf{HS Weight} & \textbf{MHS Weight} & \textbf{AP H.+State}  & \textbf{mAP H.+All} \\ \hline
			10\%               & 90\%                & 57.05                 & 35.39               \\
			20\%               & 80\%                & 57.35                 & 35.66               \\
			30\%               & 70\%                & 57.66                 & 36.08               \\
			40\%               & 60\%                & 57.96                 & 36.22               \\
			50\%               & 50\%                & \textbf{58.40}        & \textbf{36.51}      \\
			60\%               & 40\%                & \underline{58.35}     & 35.98               \\
			70\%               & 30\%                & 58.09                 & \underline{36.38}   \\
			80\%               & 20\%                & 57.89                 & 35.86               \\
			90\%               & 10\%                & 57.70                 & 35.84               \\ \hline
		\end{tabular}}
\end{table}

\subsubsection{Object Detection}\label{sec:object_detection}
We performed an additional experiment to assess the utility of using synthetic data for the related task of \textit{Object Detection}. The \textit{mean Average Precision metric}\footnote{We used the following implementation: \url{https://github.com/cocodataset/cocoapi}} with an \textit{IoU} threshold of \textit{$0.5$} (\textit{mAP@50}) was used as the evaluation criterion.
\begin{table}[t]
    \centering
    \caption{Object detection results on the \textit{EgoISM-HOI-Real} test data.}
    \label{tab:expt_object_detection}
    \resizebox{\linewidth}{!}{
    \begin{tabular}{lcc}
    \hline
    \textbf{EgoISM-HOI Synth}    & \textbf{EgoISM-HOI Real\%}    & \textbf{mAP@50\%} \\ \hline
    Yes                 & 0                     & 66.58 \\
    Yes                 & 10                    & 76.29 \\
    Yes                 & 25                    & 78.48 \\
    Yes                 & 50                    & \underline{79.68} \\
    Yes                 & 100                   & \textbf{81.06} \\ 
    No                  & 10                    & 68.41 \\
    No                  & 25                    & 71.59 \\
    No                  & 50                    & 73.33 \\
    No                  & 100                   & 72.97 \\ \hline
    \end{tabular} }
\end{table}

The results are shown in Table~\ref{tab:expt_object_detection}. The models trained using synthetic and real-world data (rows 1-5) outperform all the corresponding models trained only on the real-world training set (rows 6-9). In particular, the best result of 81.06\% was obtained by the model pre-trained on \textit{EgoISM-HOI-Synth} training set and fine-tuned with 100\% of \textit{EgoISM-HOI-Real} training data (row 5), with an improvement of +7.73\% (81.06 vs 73.33) over the model which obtains the best results among the ones trained only on \textit{EgoISM-HOI-Real} (row 8). Furthermore, it is worth noting that the model pre-trained using \textit{EgoISM-HOI-Synth} and fine-tuned with only 10\% of the \textit{EgoISM-HOI-Real} training set (row 2) surpasses all the models fine-tuned using only \textit{EgoISM-HOI-Real}.

\section{Conclusion}\label{sec:conclusion}
We studied egocentric human-object interactions in an industrial domain. Due to the expensiveness of collecting and labeling real in-domain data in the considered context, we proposed a pipeline and a tool that leverages 3D models of the objects and the considered environment to generate synthetic images of EHOIs automatically labeled and additional data signals, such as depth maps and instance segmentation masks. Exploiting our pipeline, we presented \textit{EgoISM-HOI}, a new multimodal dataset of synthetic and real EHOI images in an industrial scenario with rich annotations of hands and objects. We investigated the potential of using multimodal synthetic data to pre-train an EHOI detection system and demonstrated that our proposed method outperforms class-agnostic baselines based on the state-of-the-art method of \citet{Shan2020UnderstandingHH}. Future work will investigate how the knowledge inferred by our method can be valuable for other related tasks such as next active object detection or action recognition. Additionally, there is a need to consider the problem of handling and accurately recognizing simultaneous interactions with multiple objects. To encourage research on the topic, we publicly released the datasets and the source code of the proposed system, together with pre-trained models, on our project web page: \url{https://iplab.dmi.unict.it/egoism-hoi}.

\section*{Acknowledgments}
This research is supported by Next Vision\footnote{Next Vision: https://www.nextvisionlab.it/} s.r.l., by MISE - PON I\&C 2014-2020 - Progetto ENIGMA  - Prog n. F/190050/02/X44 – CUP: B61B19000520008, and by the project Future Artificial Intelligence Research (FAIR) – PNRR MUR Cod. PE0000013 - CUP: E63C22001940006.

\bibliographystyle{model2-names}
\bibliography{main}

\end{document}